\pdfoutput=1

\documentclass[11pt]{article}
\usepackage[table,dvipsnames]{xcolor}
\usepackage{times}
\usepackage{acl}
\usepackage{latexsym}
\usepackage[T1]{fontenc}
\usepackage[utf8]{inputenc}
\usepackage{microtype}
\usepackage{inconsolata}
\usepackage{amsmath}
\usepackage{amssymb}
\usepackage{graphicx}
\usepackage{enumerate}
\usepackage{balance}
\usepackage{blindtext}
\usepackage{listings}
\usepackage{booktabs}
\usepackage{colortbl}
\usepackage{tcolorbox}
\usepackage{marvosym}
\usepackage{float}
\usepackage{makecell}
\usepackage{multicol}
\usepackage{multirow}
\usepackage{tabularx}
\usepackage{marvosym}
\usepackage{stfloats}
\usepackage{graphicx}
\usepackage{subfigure}
\usepackage{subcaption}
\usepackage{ulem}
\usepackage{svg}
\usepackage{hyperref}
\usepackage{csquotes}
\usepackage{epigraph}
\usepackage{wrapfig}
\usepackage{amsfonts}
\usepackage{bbm}
\usepackage{algorithm}
\usepackage{wrapfig}
\usepackage{algorithmicx} 
\usepackage{algpseudocode} 
\usepackage{booktabs}
\usepackage{arydshln}

\newtheorem{definition}{Definition}

\newcommand{\eg}{\textit{e.g., }}

\definecolor{lightpurple}{RGB}{242, 242, 255}
\definecolor{lightyellow}{RGB}{252, 240, 209}

\lstdefinestyle{PythonCode}{
    language=Python,
    basicstyle=\ttfamily,
    breaklines=true,
    keywordstyle=\bfseries\color{NavyBlue},
    morekeywords={},
    emph={self},
    emphstyle=\bfseries\color{Rhodamine},
    commentstyle=\itshape\color{black!50!white},
    stringstyle=\bfseries\color{PineGreen!90!black},
    columns=flexible,
}

\lstdefinestyle{BashCode}{
    language=Bash,
    basicstyle=\ttfamily\color{white},
    backgroundcolor=\color{black},
    breaklines=true,
    keywordstyle=\bfseries\color{MidnightBlue},
    morekeywords={},
    emph={},
    emphstyle=\bfseries\color{Purple},
    commentstyle=\itshape\color{black!50!white},
    stringstyle=\bfseries\color{OliveGreen!90!black},
    columns=flexible,
}

\newcommand\footnoteONLYtext[1]{
    \let \mybackup \thefootnote
    \let \thefootnote \relax
    \footnotetext{#1}
    \let \thefootnote \mybackup
    \let \mybackup \imareallyundefinedcommand}

\title{Multi-Agent Collaboration via Cross-Team Orchestration}

\author{
\textbf{Zhuoyun Du}{\footnotesize $^{\dagger\clubsuit\natural}$} \quad
\textbf{Chen Qian}{\footnotesize $^{\dagger\diamondsuit}$} \quad
\textbf{Wei Liu}{\footnotesize $^\sharp$} \quad 
\textbf{Zihao Xie}{\footnotesize $^{\bigstar}$} \quad 
\textbf{Yifei Wang}{\footnotesize $^{\bigstar}$} \quad
\textbf{Rennai Qiu}{\footnotesize $^\bigstar$} \quad \\
\textbf{Yufan Dang}{\footnotesize $^\bigstar$} \quad
\textbf{Weize Chen}{\footnotesize $^\bigstar$} \quad 
\textbf{Cheng Yang}{\footnotesize $^\spadesuit$\textsuperscript{\Letter}} \quad 
\textbf{Ye Tian}{\footnotesize $^\flat$} \quad 
\textbf{Xuantang Xiong}{\footnotesize $^\flat$} \quad 
\textbf{Lei Han}{\footnotesize $^\flat$} \quad \\
{\footnotesize $^\clubsuit$}State Key Lab of CAD \& CG, Zhejiang University \\
{\footnotesize $^\natural$}Zhejiang Polytechnic Institute, Polytechnic Institute, Zhejiang University \quad \\
{\footnotesize $^\diamondsuit$}Shanghai Jiao Tong University \quad
{\footnotesize $^\sharp$}King's College London \quad 
{\footnotesize $^\bigstar$}Tsinghua University \quad \\
{\footnotesize $^\spadesuit$}Beijing University of Posts and Telecommunications \quad
{\footnotesize $^\flat$}Tencent Robotics X \\
\href{duzy.zju@outlook.com}{\texttt{duzy@zju.edu.cn}} \quad 
\href{qianc62@gmail.com}{\texttt{qianc@sjtu.edu.cn}} \quad
\href{yangcheng@bupt.edu.cn}{\texttt{yangcheng@bupt.edu.cn}}
}

\begin{document}

\maketitle
\footnoteONLYtext{$^\dagger$Equal Contribution.}
\footnoteONLYtext{$^{\text{\Letter}}$Corresponding Author.}

\begin{abstract}
Large Language Models (LLMs) have significantly impacted various domains, especially through organized LLM-driven autonomous agents. A representative scenario is in software development, where agents can collaborate in a team like humans, following predefined phases to complete sub-tasks sequentially. However, for an agent team, each phase yields only one possible outcome. This results in the completion of only one development chain, thereby losing the opportunity to explore multiple potential decision paths within the solution space. Consequently leading to suboptimal results or extensive trial and error. To address this, we introduce \textit{\textbf{Cro}ss-\textbf{T}eam \textbf{O}rchestration} (\textit{\textbf{Croto}}), a scalable multi-team framework that enables orchestrated teams to jointly propose various task-oriented solutions and interact with their insights in a self-independence while cross-team collaboration environment for superior solutions generation. Experiments reveal a notable increase in software quality compared to state-of-the-art baselines. We further tested our framework on story generation tasks, which demonstrated a promising generalization ability of our framework in other domains. The code and data is available at \href{https://github.com/OpenBMB/ChatDev/tree/macnet}{\texttt{https://github.com/OpenBMB/ChatDev/tr\\ee/macnet}}
\end{abstract}

\section{Introduction}
The rapid advancement of Large Language Models (LLMs) has yielded remarkable achievements across various domains like natural language processing~\cite{vaswani2017attention,brown2020language}, and software development~\cite{autogpt,dong2024self,zhang2024autocoderover}. However, limitations like hallucinations inherent in their standalone capabilities~\cite{autogpt}, impede LLM's ability to generate usable content for task solving when confronted with complexities surpassing mere chatting. A noteworthy breakthrough lies in the LLM-based collaborative autonomous agents~\cite{park2023generative,li2023camel,wu2024autogen,shinn2024reflexion}. Typical methods~\cite{qian2024communicative,hong2023metagpt} decompose tasks into several distinct sub-tasks. An instructor gives instructions and an assistant responds with a solution to solve each sub-task. Through a chained multi-turn dialog, they collaboratively generate content (\eg software, outline, scientific conclusion) for the task. The content produced can vary across multiple iterations given the same task, reflecting the dynamic nature of the problem-solving process by agents~\cite{qian2024communicative}. A series of autonomous agents interacting through multiple configurable task-oriented phases is the state-of-the-art single-team approach. The team completes the generation process through multiple sequential phases and generates task-oriented data (such as requirement documents and codes), which can be regarded as a decision path.

However, a single team can only execute all phases sequentially according to its predefined configuration (e.g., the number of agents, agent profiles, and LLM hyperparameters), and its decision path is fixed~\cite{qian2024communicative, hong2023metagpt}. This design may lead to repetitive errors with a specific configuration when encountering a particular type of problem, hindering self-correction. Furthermore, it limits the agents' ability to explore more diverse and effective decision paths. While graph-like paradigms self-organize agents through dynamic optimization of nodes and edges~\cite{zhuge2024language}, they require extensive task-specific customization for all nodes and edges. This complexity complicates their usage and hinders seamless generalization to heterogeneous downstream tasks, making them impractical in many scenarios. Additionally, organizing agents into a graph structure may reduce the task-solving independence of agents, which is crucial for fostering diverse insights into solutions.

Therefore, it is beneficial to introduce multiple agent teams that are aware of each other, enabling them to collaborate effectively to explore more potential paths without needing specific adjustments while also maintain their independence as a team process self-sufficient.
Then the challenge becomes: \textit{How can multi-agent systems obtain and utilize insights from others to achieve a superior outcome?} In this paper, we propose \textit{\textbf{Cro}ss-\textbf{T}eam \textbf{O}rchestration} (\textbf{Croto}), \textbf{a framework that carefully orchestrates different single teams into a multi-team collaborative structure}, each team has the same task assignment to communicate in a collaborative environment. 
Specifically, our framework enables different teams to independently propose various task-oriented solutions as insights (single-team proposal) and then communicate for insights interchange in some important phases (multi-team aggregation) that boost subsequent task resolution. Different solutions from various teams are divided into groups by a \textit{\textbf{Hierarchy Partitioning}} mechanism and then synthesized by a \textit{\textbf{Greedy Aggregation}} mechanism that aggregates various solutions and insights into a superior one collaboratively.

Through our experiments with 15 tasks from different categories and styles selected from the SRDD dataset~\cite{qian2024communicative} for software generation (programming-language-oriented reasoning), we demonstrate a significant improvement in software quality using the proposed framework. We highlight the importance of diversity across teams and emphasize the importance of fostering a cross-team collaboration environment to bolster teams' performance through our pruning mechanism. Furthermore, to further demonstrate the generalizability of our framework, we extended its application to the domain of story generation (natural-language-oriented reasoning), incorporating 10 tasks from the ROCStories dataset~\cite{chen2019incorporating}. The results revealed a notable improvement in story quality. Our findings underscore the efficacy and promising generalization of our framework in complex tasks.

In summary, our contributions are threefold: 
\begin{enumerate}[$\bullet$]
\item We propose Cross-Team Orchestration (Croto), a scalable multi-team collaboration framework that efficiently orchestrates agents into multiple teams to perform cross-team communications, which facilitates seamless content exchange among teams and effectively supports the generation of diverse content forms, including programming language and natural language.
\item Our approach involves concurrent reasoning within each team, followed by the partitioning and aggregation process of diverse content from multiple teams into a superior outcome, which effectively incorporates multidimensional solutions by retaining their strengths and eliminating their weaknesses.
\item We conducted extensive experiments demonstrating the effectiveness and generalizability of our framework, indicating that multi-team collaboration outperforms individual efforts.
\end{enumerate}

\section{Related Work} \label{sec:related_work}
Trained on vast datasets with extensive parameters, LLMs have revolutionized the landscape of natural language processing~\cite{brown2020language,bubeck2023sparks,vaswani2017attention,liu2024deepseek}. A notable breakthrough lies in the LLM-based autonomous agents~\cite{wang2023voyager,autogpt,gptengineer}, where these agents exhibit proficiency in planning~\cite{chen2023agentverse,yao2024tree}, memory~\cite{park2023generative,wang2024retriever}, and tool use~\cite{qin2023toolllm,yang2024gpt4tools,qin2024tool}, thus enabling independent operation within intricate real-world contexts~\cite{zhao2024expel,zhou2023webarena,zhang2023generative,LLm-powered}, thereby transforming fundamental LLMs into versatile autonomous
agents~\cite{shinn2024reflexion,zhao2024expel,lin2024swiftsage,mei2024llm}.
Along this line, multi-agent collaboration has proven beneficial in uniting the expertise of diverse agents for autonomous task-solving~\cite{khan2024debating,wang2024macrec,qian2024scaling,zhou2024symbolic}, which has widely propelled progress across various domains such as software development~\cite{qian2024communicative,hong2023metagpt}, medical treatment~\cite{tang2023medagents,li2024agent}, educational teaching~\cite{zhang2024simulating} and embodied control~\cite{chen2024scalable}.

In contrast to simple majority voting, where agents act independently~\cite{qian2024iterative}, the concept of collective emergence~\cite{woolley2010evidence,hopfield1982neural,minsky1988society} suggests that effective collaboration should form an integrated system that fosters interdependent interactions and thoughtful decision-making~\cite{li2024more,piatti2024cooperate}. Recent research has focused on differentiating agents into distinct roles and encouraging interactions for diverse tasks solving or complex  simulation~\cite{xi2025rise,li2024agent,gao2024simulating,yang2024oasis,wang2025decoding}. Studies on exploring organizing agents in hierarchical tree structures for information propagation~\cite{chen2023agentverse,wang2024mixture} or in graph-based structures with predefined nodes and edges~\cite{zhuge2024language} demonstrating that increasing the number and diversity of agents can enhance performance in multi-agent systems. MACNET~\cite{qian2024scaling} revealed that the quality of solutions follows a logistic growth pattern as the number of agents scales. Unlike these approaches, our work envisions multi-agent teams as collaborative units, enabling both inter- and intra-team collaborations for optimized solutions generation.

\section{Preliminaries} \label{sec:preliminaries}

\begin{definition}[Chain as a Team]
A single-team ($\mathcal{C}$) is conceptualized as a chain-like structure~\cite{qian2024communicative} composed of a series of task-oriented phases ($\mathcal{P}^i$) that sequentially address the resolution of tasks which can be formulated as:
\begin{equation}
\begin{aligned}
& \mathcal{C} = \langle \mathcal{P}^{1}, \mathcal{P}^{2} ,\dots,\mathcal{P}^{|\mathcal{C}|} \rangle \\
\end{aligned}
\end{equation}
We refer to such a chain-like structure as a team that could participate in our Croto framework.
\end{definition}

\begin{definition}[Agent Communication]
In each phase, an instructor agent initiates instructions, instructing ($\rightarrow$) the discourse toward the completion of the subtask, while an assistant agent adheres to these instructions and responds with ($\leadsto$) appropriate solutions~\cite{li2023camel}:
\begin{equation}
\langle \mathcal{I} \rightarrow \mathcal{A}, \quad \mathcal{A} \leadsto \mathcal{I} \rangle_{\circlearrowleft}
\end{equation}
Agents engage in a multi-turn dialogue, working collaboratively until they achieve consensus, extracting solutions that can range from the text (\eg defining a software function point) to code (\eg creating the initial version of source code), ultimately leading to the completion of the subtask.
\end{definition}

\begin{definition}[Interaction]
To facilitate intra-team collaboration for task resolution, teams propose their task-oriented solutions from the same phases and jointly participate in solution improvements. 
We refer to such a collaboration an interaction, which successfully breaks the isolation between teams while preserving their independence.
\end{definition}

\section{Methodology} \label{sec:method}

\begin{figure*}[t]
    \centering
    \includegraphics[scale=0.52]
    {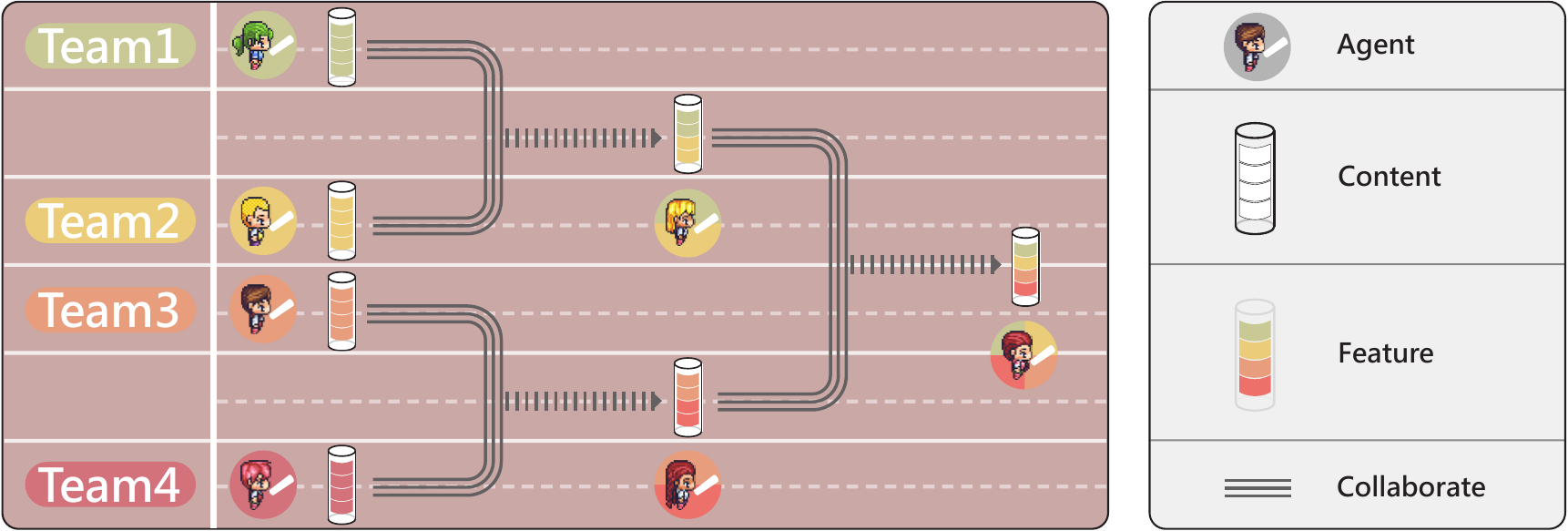}
    \caption{The aggregation process in Cross-Team Orchestration involves multiple agents (\includegraphics[height=10pt]{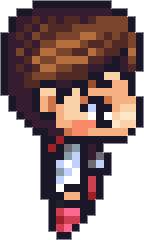}) from different teams contributing a variety of content (\includegraphics[height=10pt]{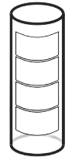}). These solutions are partitioned into groups and collaboratively (\includegraphics[height=10pt]{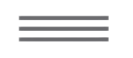}) aggregated through interactions, highlighting the distinctive features (\includegraphics[height=10pt]{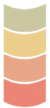}) of each team's solution. Ultimately, this process results in a superior outcome that synthesizes the features of all participating teams.}
    \label{fig:team-weaving} 
    \vspace{-10pt}
\end{figure*}

\subsection{Cross-Team Orchestration} \label{sec:weaving}
Facing diverse tasks, a chain-as-a-team often tackles tasks in isolation. While this process is streamlined, it lacks the diversity of insights beneficial to explore a broader range of decision paths for superior task solutions.

A straightforward attempt to break this isolation is to run $n$ teams simultaneously on the same task and ensemble their results. However, this straightforward ensembling approach fails to capitalize on the potential for mutual awareness and collaboration among teams during intermediate phases. This is analogous to exploring $n$ paths in parallel without leveraging the intersections of their intermediate nodes, which could enable the exploration of additional paths.
Nevertheless, this approach introduces new challenges, including a significant increase in communication overhead and the risk of incorporating noise from underperforming teams. Moreover, the lack of independence among teams can diminish the diversity of solutions, as teams may converge on similar features.

To alleviate these issues, we propose Croto, a novel cross-team collaborative framework that orchestrates the parallel executions of multiple single-team with configurable temperature and length of chains\footnote{Length Diversity can be induced manually, meanwhile, can continue to vary autonomously along the process.} Each team is assigned the same task objective, and they collectively propose various task-oriented solutions at each phase based on their unique perspectives. At predefined key phases such as design or writing, where critical decisions or significant solution modifications occur, the framework identifies corresponding key phases in other teams. If such phases exist, these teams pause their workflows and extract solutions for cross-team interactions ($\mathcal{E}$). During this interaction process, solutions generated by teams are first grouped and then iteratively aggregated into more refined solutions. This interactive dynamic can be modeled as:
\begin{equation}
\begin{aligned}
& \mathcal{N} = \{\mathcal{V}, \mathcal{E}\}, \ \ \ \ \mathcal{V} = \{v_i \mid i \in \mathcal{I}\} \\
& \mathcal{E} = \{\langle v_i, v_j \rangle \mid \mathbbm{I}(v_i) = \mathbbm{I}(v_j), v \in \mathcal{K}\}
\end{aligned}
\end{equation}

where $\mathcal{V}$ denotes the set of phases among all teams, indexed by the index set $\mathcal{I}$, $\mathbbm{I}(x)$ denotes the name of phase $x$ in a team, and $\mathcal{K}$ denotes the set of key phases. Through Cross-Team Orchestration, a collaborative yet independent cross-team interaction network is established, fostering greater innovation and efficiency in producing superior solutions.

\subsubsection{Greedy Aggregation}
During interactions, agents collaborate to jointly develop a superior solution. This process is not merely about selecting the best option but focuses on combining the strengths and mitigating deficiencies of all solutions ($\mathcal{S} = \{s_1, s_2, \dots, s_n\}$), as illustrated in Figure~\ref{fig:team-weaving}. Essentially, it synthesizes multiple decision paths into a single, optimal pathway.
To achieve this, we introduce a greedy aggregation mechanism ($\alpha$) that leverages the features of the solutions. In an aggregation process, a pruning mechanism ($\theta$) filters out a predefined proportion of low-quality solutions\footnote{Solutions are evaluated and rated using the Quality metric detailed in Section~\ref{sec:metrics}, which effectively enhances our pruning mechanism by eliminating solutions non-arbitrarily.} to reduce the aggregation burden and enhance the quality of the generated solutions. A role-assigned aggregate agent proficient in synthesizing solutions then meticulously extracts the strengths and weaknesses of each solution. Based on these features, the agent aggregates a superior ($\ast$) solution that greedily integrates strengths and eliminates weaknesses and explicitly outlines the changes that have been made:
\begin{equation}
\begin{aligned}
& s^\ast = \alpha(\theta(\mathcal{S}))
\end{aligned}
\end{equation}
The resulting solution is then disseminated to all teams, replacing prior solutions of each team, guiding subsequent phases of task resolution.

\subsubsection{Hierarchy Partitioning}
To prevent long-context issues rooted in the overwhelming amount of simultaneous solutions aggregation, meanwhile, enhance the effectiveness of the aggregation process by gradually synthesizing and refining the solutions, we propose Hierarchy Partitioning ($\tau$), as illustrated in Figure~\ref{fig:team-weaving}. This involves grouping solutions from different teams engaged in intra-team interactions and subsequently aggregating them into superior solutions by groups.

Formally, by using uniform partitioning with an expected quantity ($u$) of solutions per group, a set of collaborative groups ($\mathcal{G}^k = \{g_1^k, g_2^k, \dots, g_{\frac{n}{u^{k+1}}}^k\}$) are generated. Each group $g_i$ consists of a subset of the solutions from teams that participate in the interaction process:
\begin{equation}
\begin{aligned}
& \bigcup_{g \in \mathcal{G}} g^k_i = \mathcal{S}^k, \ \ \ \ \ \mathcal{S}^k = \{s_1^k, s_2^k, \dots, s_{\frac{n}{u^k}}^k\}
\end{aligned}
\end{equation}
where $k$ denotes the number of partitioning iterations. Following this, each group of solutions first undergoes an aggregation process, gathered and divided into new groups, and then re-aggregating these aggregated solutions. This iterative process can be formalized as:
\begin{equation}
\begin{aligned}
& \mathcal{G}^k = \tau_k(\mathcal{S}^k),\ \ \ \ \mathcal{S}^{k+1} = \alpha_k(\mathcal{G}^k) \\
& s^\ast = \alpha_x(\tau_{x}(\alpha_{x-1}(\dots \alpha_1(\tau_1(\mathcal{S}^0)))))
\end{aligned}
\end{equation}
This process continues hierarchically, generating aggregated solutions until a single, superior solution remains as the final output.

\begin{table*}[t]
\centering
\begin{tabular}{lcccccc}
\toprule[1.5pt]
\multirow{1}{*}{\textbf{Method}} & \multirow{1}{*}{\textbf{Paradigm}}& \multicolumn{1}{c}{\textbf{Completeness}} & \multicolumn{1}{c}{\textbf{Executability}} & \multicolumn{1}{c}{\textbf{Consistency} } & \multicolumn{1}{c}{\textbf{Quality}}\\
\midrule[0.75pt]
GPT-Engineer & \includegraphics[height=10pt]{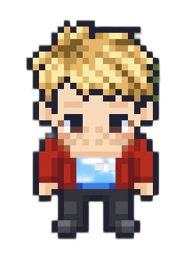}&$0.502^\dag$ &$0.358^\dag$ &$0.768^\dag$&$0.543^\dag$\\
MetaGPT & \includegraphics[height=10pt]{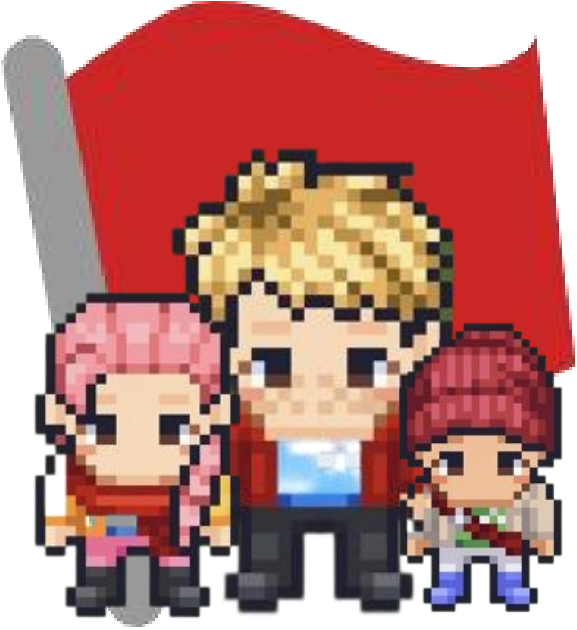}&$0.483^\dag$ &$0.415^\dag$ &$0.739^\dag$&$0.545^\dag$\\
ChatDev & \includegraphics[height=10pt]{figs/team.png}&$\underline{0.744}^\dag$ &$0.813^\dag$ &$\underline{0.781}^\dag$&$\underline{0.779}^\dag$\\
AgentVerse & \includegraphics[height=10pt]{figs/team.png}&$0.650^\dag$ &$\underline{0.850}^\dag$ &$0.776^\dag$&$0.759^\dag$\\
GPTSwarm & \includegraphics[height=10pt]{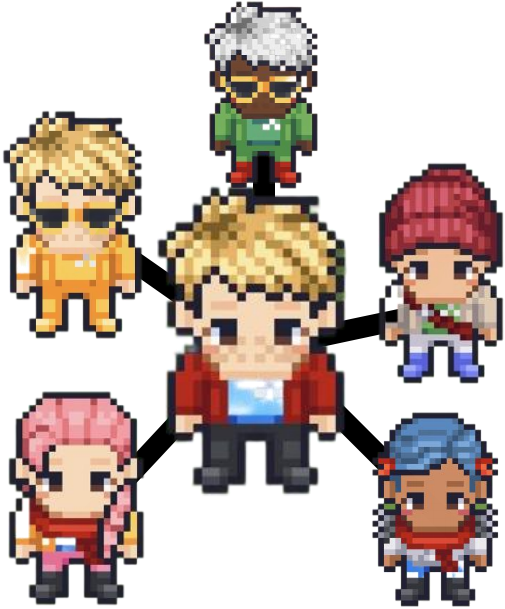}&$\textbf{0.800}$ &$0.550^\dag$ &$0.779^\dag$&$0.710^\dag$\\\midrule[0.75pt]
Croto &\includegraphics[height=10pt]{figs/team.png} \includegraphics[height=10pt]{figs/team.png}  & \underline{0.795}& \textbf{0.928} & \textbf{0.796}&\textbf{0.840}\\
\bottomrule[1.5pt]
\end{tabular}
\caption{Overall performance comparison of various representative methods, encompassing Single-Agent(\includegraphics[height=10pt]{figs/bot.png}), Single-Team Execution (\includegraphics[height=8pt]{figs/team.png}), Graph-like (\includegraphics[height=8pt]{figs/net.png}) and Our Cross Team Orchestration (\includegraphics[height=8pt]{figs/team.png} \includegraphics[height=8pt]{figs/team.png}) framework. The metrics are the average across all tasks. The highest scores are highlighted in \textbf{bold}, and the second-highest scores are presented with \underline{underline}. $\dag$ indicates significant statistical differences ($p<0.05$) between baselines and ours.}
\label{tab:main-results}
\vspace{-10pt}
\end{table*}

\section{Evaluation} \label{sec:evaluation}

\paragraph{Baselines} We chose different types of LLM-driven paradigms as our baselines, which include both single-agent and multi-agent methodologies. \textbf{GPT-Engineer}~\cite{gptengineer} is a foundational single-agent method leveraging LLMs for software development, distinguished by its adeptness at swiftly grasping task requirements and applying one-step reasoning to efficiently generate comprehensive solutions, \textbf{ChatDev}~\cite{qian2024communicative} is an LLM-powered agent collaborative software development framework that organizes the entire software development process into waterfall-style phases, \textbf{MetaGPT}~\cite{hong2023metagpt} is an innovative framework that assigns diverse roles to various LLM-powered agents and incorporates standardized operating procedures to facilitate agent collaboration in software development, \textbf{AgentVerse}~\cite{chen2023agentverse} is a multi-agent framework that assembles expert agents in structured topologies, using linguistic interactions for autonomous solution refinement, \textbf{GPTSwarm}~\cite{zhuge2024language} formalizes a swarm of LLM agents as computational graphs, where nodes represent manually customized functions and edges represent information flow.

\paragraph{Experiment Setup} 
In our experiments, we have validated our framework on heterogeneous tasks, including the scientific domain of software development and the humanities domain of story generation. 
We employ GPT-3.5-Turbo as the foundational model for its optimal balance of reasoning efficacy and efficiency. We limit communication rounds between agents to a maximum of 5 per phase in each team. By default, the number of teams engaged in the tasks is set to 8 and the temperature parameter is 0.2 with a pruning mechanism. We conduct a pruning mechanism only on 8-team Croto. We configure \textit{coding} and \textit{code completion} phases for software development tasks and after the \textit{writing} phase for story generation tasks as key phases to make cross-team interactions. Our software development experiments randomly draw 15 tasks from the SRDD dataset~\cite{qian2024communicative}, a collection designed for
repository-level software development, and 10 tasks for story generation from ROCStories~\cite{mostafazadeh2016corpus}, a collection of commonsense 5 sentences short stories can be used for longer stories generation. The performance metrics are the average across all tasks within the test set. All baseline evaluations adhere to our proposed framework's same hyperparameters and settings to ensure a fair comparison.

\paragraph{Metrics}\label{sec:metrics}
We use four fundamental dimensions to assess specific aspects of the software proposed by previous works~\cite{qian2024experiential,qian2024communicative}:

\begin{enumerate}[$\bullet$]
    \item \textit{Completeness} ($\alpha \in \left[ 0, 1 \right]$) measures the software's capacity for comprehensive code fulfillment during development. It is measured by the proportion of the software that is free from "TODO"-like placeholders. A higher score implies a greater likelihood of the software being capable of automated completion without the need for further manual coding.
    \item \textit{Executability} ($\beta \in \left[ 0, 1 \right]$) assesses the software’s ability to run correctly within a given compilation environment. It is measured by the percentage of software that compiles without errors and is ready to execute. A higher score indicates a higher likelihood of the software running successfully as intended.
    \item \textit{Consistency} ($\gamma \in \left[ 0, 1 \right]$) evaluates the alignment between the generated software and the original natural language requirements. It is quantified as the cosine distance between the embeddings of the text requirements and the source code. A higher score indicates a greater degree of compliance with the requirements.
    \item \textit{Quality} ($\frac{\alpha + \beta + \gamma}{3} \in \left[ 0, 1 \right]$) is a comprehensive metric that integrates all dimensions above. It serves as a holistic indicator of the software's overall quality. A higher score indicates superior generation quality, suggesting that the software is less likely to require additional manual interventions.
\end{enumerate}

\subsection{Overall Performance}
Table \ref{tab:main-results} illustrates a detailed comparative analysis of Croto and all baselines. Firstly, the single-team paradigm outperforms the GPT-Engineer, highlighting the benefits of a multi-agent system in decomposing complex task-solving into manageable sub-tasks. Furthermore, Croto achieved optimal performance with a remarkable improvement over baselines, showing only a slightly lower score in Completeness when compared to the Graph-like paradigm but significantly higher in Executability. The contrast with ChatDev is especially striking, the Completeness score escalates from 0.744 to 0.795, the Executability score witnesses a substantial leap from 0.813 to 0.928, the Consistency score improves from 0.781 to 0.796, the overall quality of the generated software significantly improves from 0.779 to 0.840. These enhancements underscore the advantages of the Croto, where the independence of teams maintains their solutions diversity and intra-team collaborations lead to mutual correction and enlightenment, subsequent enhancement in software quality, reducing the likelihood of executable errors, and elevating the degree of code completion and alignment with user requirements.

\subsection{Hyperparameter Analysis}
\begin{figure*}[ht]
    \centering
    \includegraphics[scale=0.25]{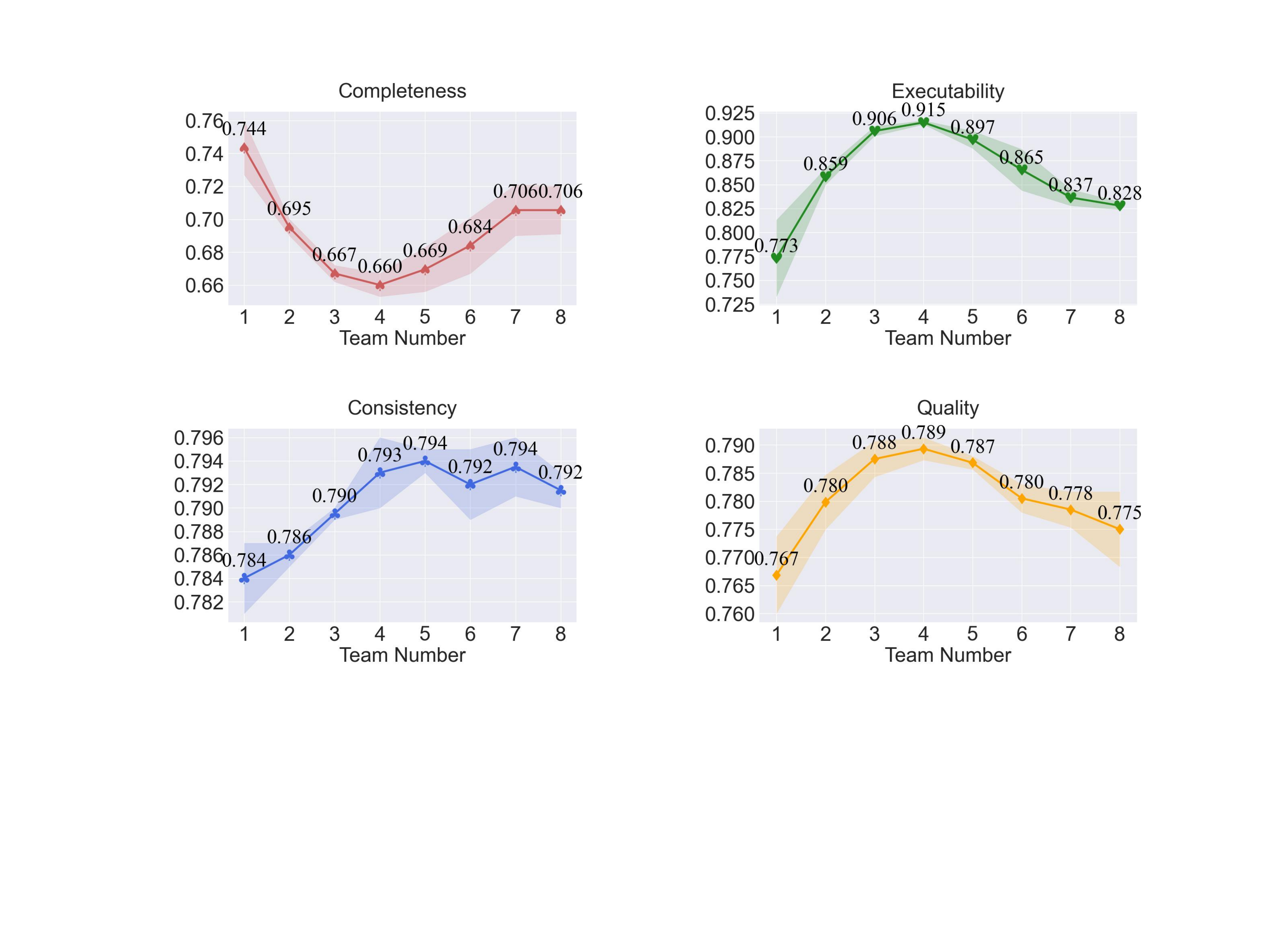}
    \caption{Visualization of result trends concerning team number variations in our framework without greedy pruning mechanism. An upward trend in consistency is observed, along with an inverse relationship between executability and completeness. The highest quality of content is achieved with a team size of four.}
    \label{fig:team_number}
    \vspace{-10pt}
\end{figure*}

\paragraph{Teams Number Analysis} Our investigation on scaling team number, as shown in Figure~\ref{fig:team_number}, reveals an intriguing inverse relationship between the Executability and Completeness of the software generated by Croto without the pruning strategy. This finding succinctly captures the essence of the trade-off that is inherent in the framework's performance. Initially, we observed a steady increase in the alignment of codes with task requirements, peaking around the 4-team configuration. This configuration achieves an optimal balance, producing software that is both executable and functionally rich. However, as the number of teams increases beyond four, we notice a gradual decline in Quality. Despite this decline, the quality remains superior to that of the single-team configuration, indicating that multi-team collaboration still offers benefits. We hypothesize that the diminishing returns and potential performance decline are due to the agents' difficulty in effectively synthesizing an excessive volume of solution features. To further scale the number of teams without compromising quality, the implementation of a pruning mechanism becomes essential, as it eliminates low-quality solutions before aggregation, effectively reducing the burden on the aggregation agent.

\begin{table}[t]
\centering
\resizebox{0.48\textwidth}{!}{
\begin{tabular}{lcccc}
\toprule[1.5pt]
\multirow{1}{*}{\textbf{Mechanism}} & \multicolumn{1}{c}{\textbf{Completeness}} & \multicolumn{1}{c}{\textbf{Executability}} & \multicolumn{1}{c}{\textbf{Consistency} } & \multicolumn{1}{c}{\textbf{Quality}}\\
\midrule[0.75pt]
\rowcolor{lightyellow}8-team Croto & $\underline{0.706}^\dag$ & $\underline{0.828}^\dag$& $\underline{0.792}^\dag$& $\underline{0.775}^\dag$\\
\rowcolor{lightpurple}+ Prune & \textbf{0.795}& \textbf{0.928} & \textbf{0.796}&\textbf{0.840}\\
\hdashline
\rowcolor{lightpurple}$\Delta$ compared to Vanilla & \color{red!80!black}+0.089 & \color{red!80!black}+0.100 & \color{red!80!black}+0.004 & \color{red!80!black}+0.065\\
\bottomrule[1.5pt]
\rowcolor{lightyellow}4-team Croto & 0.660 & $\underline{0.915}^\dag$&\textbf{0.793}& $\underline{0.789}^\dag$\\
(0.1 0.1 0.1 0.1) &\textbf{0.700}&$0.794^\dag$&0.791&$0.762^\dag$\\
(0.4 0.4 0.4 0.4) &0.583&$0.773^\dag$ & \underline{0.792}& $0.716^\dag$\\
(0.2 0.4 0.6 0.8) &0.575&$0.875^\dag$ & 0.790& $0.747^\dag$\\
\rowcolor{lightpurple}(0.2 0.2 0.4 0.4) &\underline{0.670}& \textbf{0.925} &0.790& \textbf{0.795}\\
\hdashline
\rowcolor{lightpurple}$\Delta$ compared to Vanilla & \color{red!80!black}+0.010 & \color{red!80!black}+0.010 & \color{red!80!black}+0.003 & \color{red!80!black}+0.006\\
\midrule[1.50pt]
\end{tabular}
}
\caption{Investigation of mechanisms in 4-team and 8-team Croto. The temperatures for each team are indicated as $(t_1, t_2, t_3, t_4)$. The '+' symbol represents the adding operation. 
}
\label{tab:tem-prune}
\vspace{-10pt}
\end{table}

\paragraph{Greedy Pruning} 
To enhance the scalability and performance of Croto, we introduce the Greedy Pruning mechanism, which reduces the aggregation burden and improves the quality of solutions generated by teams. As shown in Table~\ref{tab:tem-prune}, applying pruning in the 8-team configuration achieves the highest scores across all metrics, demonstrating its effectiveness in handling larger team sizes. By evaluating solutions before aggregation, Greedy Pruning eliminates low-quality solutions that could otherwise degrade the final output by introducing suboptimal features and increasing the aggregation burden~\footnote{Burden in the sense that the agent is instructed to synthesize features from all solutions in a group into one solution; more solutions increase the difficulty of aggregation.}. This makes our framework more scalable and effective for software development tasks that support Croto in exploring more valuable pathways in a cost-efficient manner while reducing the likelihood of being misled by failed paths.

\begin{table}[htbp]
\centering
\resizebox{0.48\textwidth}{!}{
\begin{tabular}{lcccc}
\toprule[1.5pt]
\multirow{1}{*}{\textbf{Mechanism}} & \multicolumn{1}{c}{\textbf{Completeness}} & \multicolumn{1}{c}{\textbf{Executability}} & \multicolumn{1}{c}{\textbf{Consistency} } & \multicolumn{1}{c}{\textbf{Quality}}\\
\midrule[0.75pt]
\rowcolor{lightpurple}4-team Croto &0.660 & \textbf{0.915}&\textbf{0.793}&\textbf{0.789}\\
- Partition &\color{red!80!black}\textbf{0.683} & \color{green!50!black}\underline{0.800}&\color{green!50!black}\underline{0.786}&\color{green!50!black}0.756\\
- Role &\color{red!80!black}\underline{0.680}&\color{green!50!black}0.783 &\color{green!50!black}0.739&\color{green!50!black}\underline{0.735}\\\midrule[1.50pt]
\rowcolor{lightpurple}8-team Croto &\underline{0.706} & \textbf{0.828}&\textbf{0.791}&\textbf{0.775}\\
- Partition &\color{red!80!black}\textbf{0.728} & \color{green!50!black}\underline{0.804}&\color{green!50!black}0.787&\color{green!50!black}\underline{0.773}\\
- Role &\color{green!50!black}0.658&\color{green!50!black}0.783 &\color{green!50!black}\underline{0.790}&\color{green!50!black}0.744\\
\bottomrule[1.5pt]
\end{tabular}
}
\caption{Ablation study on 4 Teams Croto and 8 Teams Croto. The '-' denotes the removing operation. The highest scores are highlighted in \textbf{bold}, and the second-highest scores are presented with \underline{underline}.
}
\label{tab:ablation}
\vspace{-15pt}
\end{table}

\paragraph{Temperature Analysis}
A core idea of our framework is that diverse solutions from multiple teams provide valuable perspectives that, while individually inconspicuous, can be synthesized to positively contribute to task resolution. To evaluate the impact of solution diversity, we varied temperature configurations to enable teams to generate solutions with varying degrees of creativity and requirement compliance. As shown in Table~\ref{tab:tem-prune}, an optimal level of diversity significantly improves solution quality. When the temperature is uniform across all teams, the performance gains achieved by Croto are limited. This limitation arises because teams either uniformly prioritize creative but unstable solutions or strictly adhere to rules, resulting in minimal novel insights from cross-team interactions. In contrast, when each team's temperature is set to balance creativity and compliance, Croto demonstrates substantial performance improvements. Analysis of team solutions reveals that cross-team communication often leads to autonomous functional enhancements (e.g., innovative GUI designs, progressively increasing game difficulty), facilitating the integration of beneficial features from diverse solutions.

\begin{table}[t]
\centering
\resizebox{0.48\textwidth}{!}{
\begin{tabular}{lcccc}
\toprule[1.5pt]
\multirow{1}{*}{\textbf{Mechanism}} & \multicolumn{1}{c}{\textbf{\#Token}} & \multicolumn{1}{c}{\textbf{\#Files}} & \multicolumn{1}{c}{\textbf{\#Lines}} & \multicolumn{1}{c}{\textbf{Duration (s)}}\\\midrule[0.75pt]
\rowcolor{lightyellow}Single Team & 24377 & 3.13 & 104.6 & 164.36 \\
2-team Croto & 32963 & 3.23 & 113.4 & 308.29 \\
3-team Croto & 34896 & 3.85 & 124.1 & 532.53 \\
4-team Croto & 41903 & 4.46 & 135.3 & 418.34 \\
5-team Croto & 44987 & 4.31 & 128.5 & 433.44 \\
6-team Croto & 45578 & 3.95 & 128.4 & 461.56 \\
7-team Croto & 48812 & 4.37 & 126.2 & 427.08 \\
\rowcolor{lightpurple}8-team Croto & 52179 & 4.77 & 129.6 & 584.83 \\
\hdashline
\rowcolor{lightpurple}$\Delta$ compared to single & $\times$2.141 & $\times$1.524 & $\times$1.239 & $\times$3.558 \\
\bottomrule[1.5pt]
\end{tabular}
}
\caption{Software statistics include Duration (time consumed), \#Tokens (number of tokens used), and \#Lines (total lines of code per across all files).}
\label{tab:cost}
\vspace{-10pt}
\end{table}

\subsection{Ablation Study}
In our ablation study, as presented in Table~\ref{tab:ablation}, the removal of Hierarchical Partitioning from the 4-team and 8-team configurations significantly reduced solution quality, from 0.789 and 0.775 to 0.756 and 0.773, respectively. This indicates that, without partitioning into groups, the aggregate agent struggled to effectively handle the diverse features of team solutions in a single aggregation, making it challenging or even impractical to extract and synthesize these features. Furthermore, eliminating role assignment for the aggregate agent further decreased solution quality to 0.735 and 0.744. The absence of structured guidance led to issues such as disorganized solutions, task failures, and feature omissions. These results underscore the importance of our framework's mechanisms in managing complex solutions and ensuring high-quality outputs in multi-team scenarios.

\subsection{Statistics Analysis}
We present the software statistics in Table~\ref{tab:cost}. Croto generates a greater number of code files and a larger codebase, significantly enhancing the software’s functionality and integrity. This trend is consistent across configurations ranging from 2 to 8 teams, demonstrating the effectiveness and scalability of our framework. The increased number of files reflects a more structured programming architecture, resembling software developed by a sophisticated software development team. Although slower and more token-intensive than the single-agent method, our framework remains computationally efficient, as the duration and token consumption do not scale linearly. Specifically, these metrics increase only 2.14 times more tokens and 3.558 times more duration when scaling from a single team to 8 teams. This efficiency is attributed to solution elimination through greedy pruning and the fact that higher-quality aggregated solutions reduce processing complexity in subsequent phases, resulting in fewer average communication rounds per phase and lower token consumption. 
Considering these factors, we posit that the fundamental characteristics of cross-team software development hold greater significance, outweighing short-term concerns such as time and economic costs in the current landscape.

\subsection{Diversity in Collaborative Emergence}
\label{subsec:diversity}
The diversity in Croto arises from the interaction among teams, as formalized by the equation:
\begin{equation}
    \begin{aligned}
    & p^n(t) = 1 - (1 - p(t))^n \\
    & \quad \quad \quad \quad \quad \quad \propto 1 - (1 - 1/r(t))^{|V|^2}, \\
    & \lim_{|V|\rightarrow \infty}p^n(t) = \lim_{n\rightarrow\infty}p^n(t) = 1.
    \end{aligned}
\label{eq:diversity}
\end{equation}
Here, increasing the number of teams ($|V|$) quadratically enhances the likelihood of capturing rare but valuable \textit{long-tail} solution features---such as unconventional yet effective code logic or creative narrative twists---since token distributions in underlying models typically follow a long-tail pattern. This differs from conventional linear agent-level optimization, where rare features are less systematically integrated. The probability $p(t)$ of a rare feature appearing follows a long-tail distribution consistent with Zipf's law~\cite{newman2005power}, such that $p(t) \propto 1/r(t)$, where $r(t)$ denotes the frequency rank of feature $t$. The sampling size $n$ is proportional to team interaction density ($n \propto |V|^2$), as solution features are aggregated and refined through inter-team comparisons. Consequently, $p^n(t)$, the probability of observing feature $t$ at least once, grows quadratically with $|V|$. As the number of teams increases, the emergence of rare features becomes inevitable, enabling the aggregation process to refine solutions with increasingly nuanced aspects. This mechanism exploits multi-agent interaction scaling to improve solution diversity and quality, aligning with findings in multi-agent debate and cross-examination frameworks~\cite{liang2023encouraging,du2023improving,cohen2023lm}.

\subsection{Generalizability Analysis}
\begin{table*}[t]
\centering
\resizebox{0.95\textwidth}{!}{
\begin{tabular}{lcccccc}
\toprule[1.5pt]
\multirow{1}{*}{\textbf{Mechanism}} & \multirow{1}{*}{\textbf{Paradigm}}& \multicolumn{1}{c}{\textbf{Grammar and Fluency}} & \multicolumn{1}{c}{\textbf{Context Relevance}} & \multicolumn{1}{c}{\textbf{Logic Consistency} } & \multicolumn{1}{c}{\textbf{Quality}}\\
\midrule[0.75pt]
Single-Agent & \includegraphics[height=10pt]{figs/bot.png}&$2.150^\dag$ &$2.005^\dag$ &$2.425^\dag$&$2.193^\dag$\\
Single-Team Execution & \includegraphics[height=10pt]{figs/team.png}&$2.250^\dag$ &$2.325^\dag$ &$2.500^\dag$&$2.358^\dag$\\\midrule[0.75pt]
2-team Croto &\includegraphics[height=10pt]{figs/team.png} \includegraphics[height=10pt]{figs/team.png}  &2.725&2.800&3.000&2.842\\
3-team Croto &\includegraphics[height=10pt]{figs/team.png} \includegraphics[height=10pt]{figs/team.png}  &2.967&2.767&2.967&2.900\\
4-team Croto &\includegraphics[height=10pt]{figs/team.png} \includegraphics[height=10pt]{figs/team.png}  &2.967&2.850 &2.908&2.908\\
5-team Croto &\includegraphics[height=10pt]{figs/team.png} \includegraphics[height=10pt]{figs/team.png}  &2.980&2.880&2.960&2.940\\
6-team Croto &\includegraphics[height=10pt]{figs/team.png} \includegraphics[height=10pt]{figs/team.png}  &2.983&2.900 &2.983&2.956\\
7-team Croto &\includegraphics[height=10pt]{figs/team.png} \includegraphics[height=10pt]{figs/team.png}  &\underline{3.000}&3.171&\underline{3.014}&3.062\\
8-team Croto &\includegraphics[height=10pt]{figs/team.png} \includegraphics[height=10pt]{figs/team.png}  
&$\underline{3.000}^\dag$&$\underline{3.250}^\dag$ &$3.000^\dag$&$\underline{3.083}^\dag$\\\midrule[0.75pt]
8-team Croto + Prune &\includegraphics[height=10pt]{figs/team.png} \includegraphics[height=10pt]{figs/team.png}\includegraphics[height=10pt]{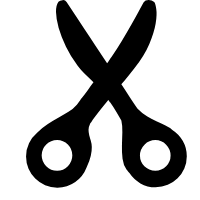}
& \textbf{3.625}& \textbf{3.750}& \textbf{3.250}& \textbf{3.642}\\
\bottomrule[1.5pt]
\end{tabular}
}
\caption{Result trends concerning Team Size Variations in our Framework in Story Generation, encompassing single-agent(\includegraphics[height=10pt]{figs/bot.png}), Single-Team Execution (\includegraphics[height=10pt]{figs/team.png}) and Cross-Team Orchestration (\includegraphics[height=10pt]{figs/team.png} \includegraphics[height=10pt]{figs/team.png}) with and without pruning mechanism (\includegraphics[height=10pt]{figs/cut.png}). $\dag$ indicates significant statistical differences ($p<0.05$) between best results and baselines}
\label{tab:story-results}
\vspace{-10pt}
\end{table*}

To demonstrate the generalization capability of our framework, we conducted experiments in story generation. The results indicate that our framework significantly enhances the quality of stories generated by both single-agent and single-team execution. This improvement highlights the versatility, robustness, and potential of our framework across diverse domains.

\paragraph{Metrics}
We evaluate story quality across four critical dimensions:

\begin{enumerate}[$\bullet$]
    \item \textbf{\textit{Grammar and Fluency}} ($\omega \in \left[ 0, 4 \right]$) assesses natural language use, grammatical correctness, and fluency for a coherent and error-free narrative flow.
    \item \textbf{\textit{Context Relevance}} ($\psi \in \left[ 0, 4 \right]$) analyzes the contextual appropriateness and interrelation of names, pronouns, and phrases to ensure narrative integrity and depth in plots.
    \item \textbf{\textit{Logic Consistency}} ($\xi \in \left[ 0, 4 \right]$) examines the logical progression of events and character relationships for narrative coherence and plausibility.
    \item \textbf{\textit{Quality}} ($\frac{\omega + \psi + \xi}{3} \in \left[ 0, 4 \right]$) is a comprehensive metric that integrates individual dimension scores to provide a comprehensive measure of narrative quality, reflecting the synthesis of language, context, and logic.
\end{enumerate}

\paragraph{Team Number Analysis} 
Experiments on team number shown in Table~\ref{tab:story-results}, observe a positive correlation between the number of participating teams and the resultant quality of the generated stories. Notably, the quality demonstrated a substantial improvement over outcomes from single agent and single-team baselines, with scores rising from 2.193 and 2.358 to 3.083. However, as the number of teams increased, diminishing returns began to set in. To counteract this trend, we bring in the Greedy Pruning mechanism. This intervention led to a notable enhancement in quality when the number of teams was eight, with the quality score improving from 3.083 to 3.642. These findings underscore the efficacy of the Croto framework in story generation, suggesting that it is not only beneficial for software development but also generalizes well to creative humanities domains such as narrative generation.

\paragraph{Ablation Study}
Our ablation study, as illustrated in Table~\ref{tab:ablation}, reveals results that align with patterns observed in software development. The removal of partitioning from the 4-team and 8-team configurations resulted in a decline in quality scores, from 2.908 and 3.083 to 2.271 and 2.456, respectively. Similarly, eliminating role assignments for agents further reduced the quality scores from 2.908 and 3.083 to 2.300 and 2.341.
While story generation exhibits more literary characteristics compared to software development tasks—which demand precision and error-free execution—they possess a higher tolerance for ambiguity. Unlike software features (\eg GUI, object-oriented programming), stories encompass more implicit features (\eg narrative style and thematic intent). These features necessitate that aggregation agents, equipped with assigned roles, process a manageable volume of stories to effectively extract and harmonize these features, thereby producing higher-quality narratives.
Given these similarities between the two distinct types of tasks, we hypothesize that content generation tasks (\eg code, stories, reports, blogs) may similarly benefit from our framework, underscoring its potential broad applicability.

\begin{table}[htbp]
\centering
\resizebox{0.48\textwidth}{!}{
\begin{tabular}{lcccl}
\toprule[1.5pt]
\multirow{2}{*}{\textbf{Mechanism}} & \multicolumn{1}{c}{\textbf{Grammar}} & \multicolumn{1}{c}{\textbf{Context}} & \multicolumn{1}{c}{\textbf{Logic}} & \multirow{2}{*}{\textbf{Quality}}\\
& \multicolumn{1}{c}{\textbf{Fluency}} & \multicolumn{1}{c}{\textbf{Relevance}} & \multicolumn{1}{c}{\textbf{Consis.} } & \\
\midrule[0.75pt]
\rowcolor{lightpurple}4-team Croto &\textbf{2.967}&\textbf{2.850} &\textbf{2.908}&\textbf{2.908}\\
- Partition &\color{green!50!black}1,906 & \color{green!50!black}\underline{2.219}&\color{green!50!black}\underline{2.688}&\color{green!50!black}2.271\\
- Role &\color{green!50!black}\underline{2.096}&\color{green!50!black}2.183 &\color{green!50!black}2.621&\color{green!50!black}\underline{2.300}\\\midrule[1.50pt]
\rowcolor{lightpurple}8-team Croto & \textbf{3.000}& \textbf{3.250}& \textbf{3.000}& \textbf{3.083}\\
- Partition &\color{green!50!black}\underline{2.255} & \color{green!50!black}\underline{2.354}&\color{green!50!black}\underline{2.758}&\color{green!50!black}\underline{2.456}\\
- Role &\color{green!50!black}2.115&\color{green!50!black}2.256&\color{green!50!black}2.653&\color{green!50!black}2.341\\
\bottomrule[1.5pt]
\end{tabular}
}
\caption{Ablation study on 4 Teams Croto and 8 Teams Croto. The '-' denotes the removing operation. 
The highest scores are highlighted in \textbf{bold}, and the second-highest scores are presented with \underline{underline}.
}
\label{tab:ablation_story}
\vspace{-10pt}
\end{table}

\section{Conclusion} \label{sec:conclusion}
Recognizing the inherent limitations of a single team in obtaining and leveraging external insights when completing complex tasks such as software development, we introduce a novel multi-team framework called Croto. This framework carefully orchestrates multiple teams with the same task objective, enabling them to jointly propose diverse task-oriented decisions, interact at key phases, and collaboratively aggregate various solutions into a final superior outcome. Without requiring task-specific customization, our quantitative analysis demonstrates significant improvements in outcome quality in software development and story generation, highlighting the framework's scalability and potential generalizability. We anticipate that our insights will initiate a paradigm shift in the design of LLM agents, advancing them toward multi-team collaboration and enhancing solution generation quality across a broader range of complex tasks.

\newpage

\section{Limitations} \label{sec:limitations}
Our study has explored the collaborative behaviors of multiple autonomous agent teams in software development and story generation, yet both researchers and practitioners must be mindful of certain limitations and risks when using the approach to develop new techniques or applications.

Firstly, the framework's dependence on a greedy pruning mechanism could inadvertently lead to the discarding of potentially valuable insights. This is due to the imperfections inherent in evaluation metrics. While the mechanism aims to eliminate low-quality solutions, it may also prematurely exclude creative solutions that could evolve into high-quality outcomes with further development. There is a trade-off between the efficiency of the pruning process and the potential loss of innovative ideas, which suggests the need for more effective automated evaluation methods in the future, not limited to the domains of software development and story generation.

Secondly, when evaluating the capabilities of autonomous agents from a software development standpoint, it is prudent to avoid overestimating their software production abilities. Our observations indicate that while Cross-Team Orchestration (Croto) significantly improves the quality of both software development and story generation tasks, autonomous agents often default to implementing the most straightforward logic during the software creation process. In the absence of explicit and clear requirements, agents struggle to autonomously discern the underlying concepts and nuances of the task requirements. For example, when developing a Flappy Bird game, if the task guidelines are not meticulously defined, agents may default to representing the bird and tubes with a rudimentary rectangular shape. Similarly, in the construction of an information management system, agents may opt to hard-code the information to be queried in a basic key-value format directly into the code, rather than employing a more sophisticated and flexible external database solution. Therefore, we advocate for the precise definition of detailed software requirements. This includes specifying whether a user interface is essential, if there is a need for the automatic generation of game character assets, or if an external database is necessary. Given the current capabilities of autonomous agents, fulfilling highly detailed requirements is not always assured, underscoring the importance of striking a balance between specificity and practical feasibility in the requirements. In the field of story generation, due to its literary nature, complex task relationships, scene descriptions, and background settings are often required. However, providing agents with overly complex requirements can lead to suboptimal narrative outcomes, as agents may find it challenging to effectively manage and prioritize the various narrative elements during the writing process. In conclusion, the research on autonomous agents for software and story generation is still in its early stages, and the associated technologies are not yet readily adaptable to complex real-world scenarios. As a result, the current application of these technologies is more suited to the development of prototype systems rather than fully-fledged, real-world software and narrative systems.

Thirdly, the complexity of coordinating multiple teams and managing the interaction load increases with the number of teams involved. As the framework scales, the computational and logistical demands rise, which may impact the practicality of applying our framework to very large-scale problems or in resource-constrained environments. Future work is needed to optimize the scalability of the framework while maintaining its efficacy.

\normalem

\bibliography{ref}

\end{document}